\documentclass{article}

\usepackage{microtype,graphicx,subcaption,booktabs,hyperref,amsmath,amssymb,mathtools,amsthm,float,multicol,multirow,subcaption,colortbl}

\usepackage[preprint]{icml2026}

\usepackage[capitalize,noabbrev]{cleveref}

\theoremstyle{plain}

\theoremstyle{definition}

\theoremstyle{remark}

\usepackage[textsize=tiny]{todonotes}
\usepackage{enumitem}

\icmltitlerunning{Boosting SAM for Cross-Domain Few-Shot Segmentation via Conditional Point Sparsification}

\begin{document}

\twocolumn[
  \icmltitle{Boosting SAM for Cross-Domain Few-Shot Segmentation\\via Conditional Point Sparsification}

  \icmlsetsymbol{equal}{*}
  \icmlsetsymbol{corresponding}{$\dagger$}

  \begin{icmlauthorlist}
    \icmlauthor{Jiahao Nie}{equal,ntu,teleai}
    \icmlauthor{Yun Xing}{equal,ntu}
    \icmlauthor{Wenbin An}{xjtu}
    \icmlauthor{Qingsong Zhao}{teleai}
    \icmlauthor{Jiawei Shao}{teleai}\\
    \icmlauthor{Yap-Peng Tan}{vinuni,ntu}
    \icmlauthor{Alex C. Kot}{ntu}
    \icmlauthor{Shijian Lu}{ntu}
    \icmlauthor{Xuelong Li}{corresponding,teleai}
  \end{icmlauthorlist}

  \icmlaffiliation{ntu}{Nanyang Technological University}
  \icmlaffiliation{teleai}{Institute of Artificial Intelligence (TeleAI), China Telecom}
  \icmlaffiliation{xjtu}{Xi'an Jiaotong University}
  \icmlaffiliation{vinuni}{VinUniversity}

  \icmlcorrespondingauthor{Jiahao Nie}{jiahao007@e.ntu.edu.sg}
  \icmlcorrespondingauthor{Xuelong Li}{xuelong\_li@ieee.org}

  \icmlkeywords{Cross-Domain Few-Shot Segmentation}

  \vskip 0.3in
]

% Use ONE of the following lines. DO NOT remove the command.
% If you have no special notice, KEEP empty braces:
% \printAffiliationsAndNotice{}  % no special notice (required even if empty)
% Or, if applicable, use the standard equal contribution text:
\printAffiliationsAndNotice{\hspace{-4.4mm} \icmlEqualContribution \ \ $^\dagger$\text{Corresponding author}\\}

\newcommand{\blue}[1]{{\textit{\textcolor{blue}{#1}}}}

\definecolor{mycolor}{rgb}{0.886, 0.949, 0.996}

\begin{abstract}

Motivated by the success of the Segment Anything Model (SAM) in promptable segmentation, recent studies leverage SAM to develop training-free solutions for few-shot segmentation, which aims to predict object masks in the target image based on a few reference exemplars. These SAM-based methods typically rely on point matching between reference and target images and use the matched dense points as prompts for mask prediction. However, we observe that dense points perform poorly in Cross-Domain Few-Shot Segmentation (CD-FSS), where target images are from the medical or satellite domain. We attribute this issue to large domain shifts that disrupt the point–image interactions learned by SAM, and find that point density plays a crucial role under such conditions. To address this challenge, we propose Conditional Point Sparsification (CPS), a training-free approach that adaptively guides SAM interactions for cross-domain images based on reference exemplars. Leveraging ground-truth masks, the reference images provide reliable guidance for adaptively sparsifying dense matched points, enabling more accurate segmentation results. Extensive experiments demonstrate that CPS outperforms existing training-free SAM-based methods across diverse CD-FSS datasets.

\end{abstract}

\section{Introduction}\label{sec:introduction}

\begin{figure}[t]
    \centering
    \vspace{-4mm}
    \includegraphics[width=\linewidth]{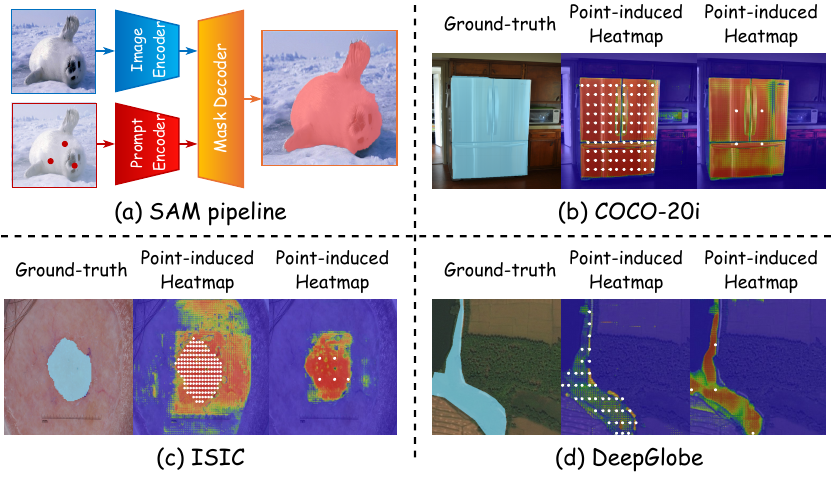}
    \vspace{-6.5mm}
    \caption{(a) Overview of SAM for promptable segmentation. (b–d) Existing methods exhibit inconsistent segmentation behavior between in-domain and cross-domain images.}
    \label{fig:motivation}
    \vspace{-3mm}
\end{figure}

Trained on billions of prompt-image-mask triplets, the Segment Anything Model (SAM)~\cite{kirillov2023segment} demonstrates unprecedented segmentation capability across diverse objects and granularities. Its promptable segmentation capability has enabled a wide range of applications, including automated mask annotation~\cite{zhang2025alps,wang2023sammed}, robotics~\cite{fang2025sam2act,pan2025omnimanip}, and virtual reality~\cite{yang2025sam}. Motivated by SAM’s strong ability to segment prompt-specified objects using prompt points (see Fig.~\ref{fig:motivation}(a)), recent works~\cite{liu2023matcher,zhang2024bridge} have explored training-free SAM-based methods for Few-Shot Segmentation, a task that aims to segment target objects conditioned on a few reference exemplars~\cite{wang2019panet,zhang2019canet}. These methods typically locate prompt points in the target images via dense matching with the reference images and then use these points as SAM prompts to segment the target objects.

Although existing methods perform well on in-domain datasets (\textit{e.g.}, COCO-20i~\cite{nguyen2019feature})~\cite{liu2023matcher}, we observe significant performance degradation when they are directly applied to Cross-Domain Few-Shot Segmentation (CD-FSS) datasets, such as medical~\cite{codella2019skin,tschandl2018ham10000} and satellite images~\cite{demir2018deepglobe}. As shown in Fig.~\ref{fig:motivation}(b–d), even when the dense prompt points consistently fall within the target regions (\textit{i.e.}, \textit{refrigerator}, \textit{skin nevus}, and \textit{water area}), the segmentations for \textit{skin nevus} and \textit{water area} remain noticeably less accurate. More illustrative examples are provided in the appendix.

\begin{figure}[t]
    \centering
    \vspace{-1mm}
    \includegraphics[width=0.97\linewidth]{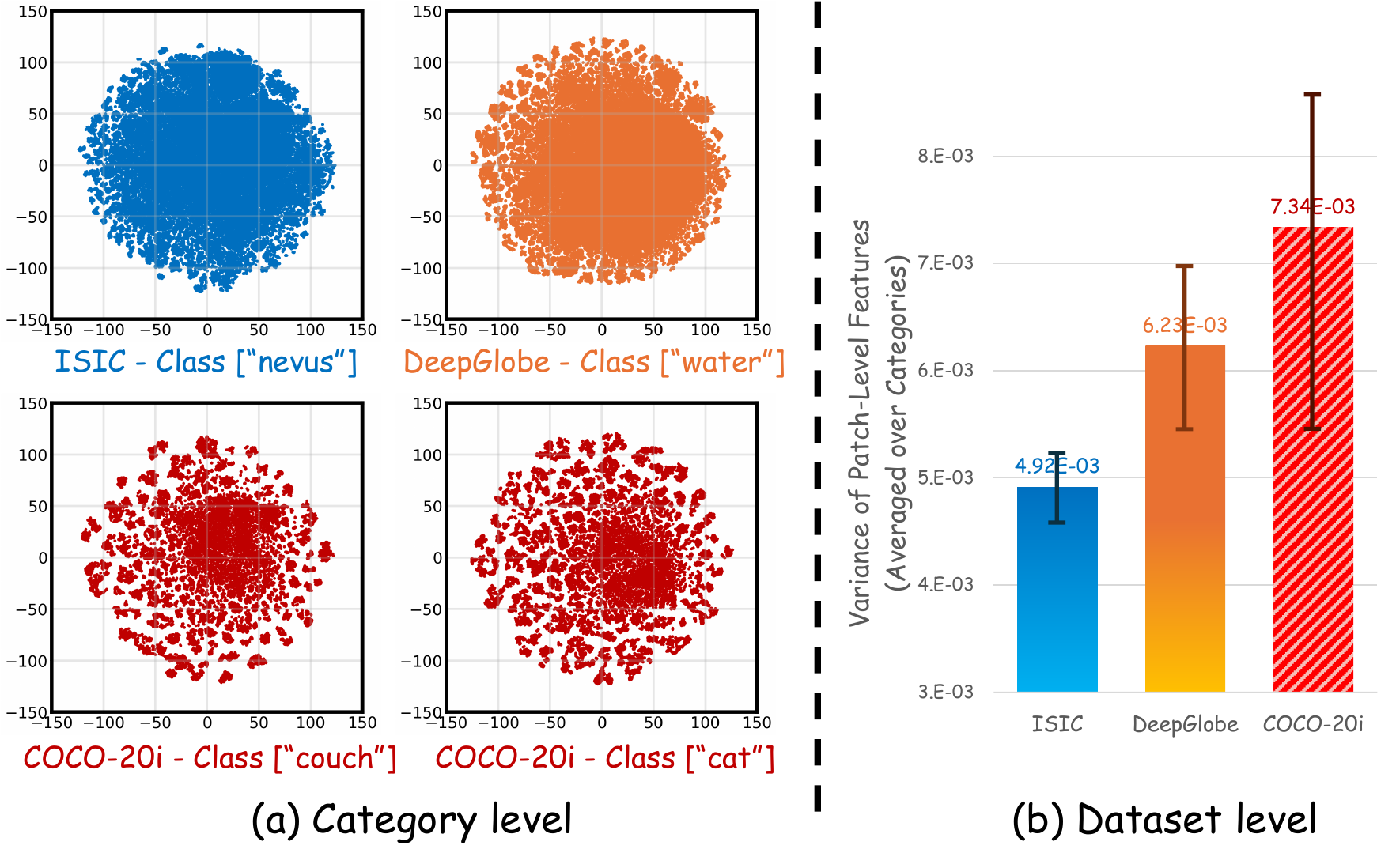}
    \vspace{-2mm}
    \caption{Analysis of SAM-encoded target-object patch features. (a) Category level: t-SNE visualization shows that in-domain images exhibit larger intra-category variance than cross-domain images. (b) Dataset level: Mean variance of all categories (bars) and inter-category variance range (error bars) confirm this trend.}
    \vspace{-3mm}
    \label{fig:tsne}
\end{figure}

We attribute these phenomena to significant domain shifts that disrupt the learned point–image interactions in SAM. Fig.~\ref{fig:tsne} shows t-SNE~\cite{maaten2008tsne} visualizations of target object features extracted by the SAM vision encoder, revealing clear distribution differences across datasets. We observe that the intra-object feature variance in medical and satellite images is considerably lower than that in COCO-20i. This is consistent with dataset characteristics: objects such as \textit{skin nevus} and \textit{water area} exhibit more uniform and homogeneous semantics, whereas an object like \textit{cat} contains diverse parts (\textit{e.g.}, \textit{head}, \textit{leg}, and \textit{tail}). Moreover, objects in in-domain datasets typically have more complex shapes~\cite{lang2023base}, for which dense points are used during SAM training to provide precise spatial guidance. However, when transferred to cross-domain scenarios, such dense points become less suitable and introduce ambiguity: the more homogeneous semantics of these objects require fewer but more informative points. As illustrated in Fig.~\ref{fig:motivation}(b–d), sparse points can therefore produce more accurate region activations in these cases. Additional examples are provided in the appendix.

To address the above limitations and extend SAM-based solutions to CD-FSS, we propose Conditional Point Sparsification (CPS), a novel training-free framework that leverages reference exemplars to guide prompt points in the target image. Our approach is grounded in a key observation (Fig.~\ref{fig:motivation}(c-d)) that sparse points lead to more reliable SAM interactions, whereas the dense prompts used in existing methods often degrade performance in cross-domain scenarios. Specifically, CPS infers an appropriate point density from few-shot reference exemplars with ground-truth masks, and utilizes this reference density to sparsify the matched points. This conditional sparsification strategy aligns point density with dataset characteristics, thereby improving point–image interactions under domain shifts. Despite its simplicity, CPS consistently outperforms prior SAM-based FSS methods (\textit{e.g.}, Matcher~\cite{liu2023matcher}) across four cross-domain benchmarks with only lightweight modifications, and also surpasses heuristic sparsification schemes based on manually defined hyperparameters.

The contributions of this work can be summarized in three aspects: \textbf{\textit{First}}, we reveal that the performance degradation of existing SAM-based FSS methods in cross-domain images stems from domain shifts that break the learned point–image interactions in SAM. \textbf{\textit{Second}}, we propose Conditional Point Sparsification (CPS), a simple yet effective training-free strategy that infers appropriate prompt density from few-shot reference exemplars and adopts it to sparsify the matched points in target images. \textbf{\textit{Third}}, CPS improves performance over prior SAM-based training-free approaches across challenging cross-domain benchmarks.
\vspace{-2mm}
\section{Related Work}\label{sec:related_work}

\noindent\textbf{Cross-Domain Few-Shot Segmentation (CD-FSS)} aims to segment target objects conditioned on a few reference exemplars under significant domain shifts~\cite{lu2021simpler,boudiaf2021few,wang2022remember,tong2025self,wu2024task,chen2024cross,chen2024pixel}. Early works~\cite{lei2022cross,fu2024cross,nie2024cross} typically adopt a two-stage paradigm, where an ImageNet~\cite{deng2009imagenet} pre-trained model is first trained on a large-scale source domain and then adapted to each target domain. These methods explore dynamic adaptation, refinement~\cite{fan2023darnet}, knowledge transfer~\cite{huang2023restnet}, feature frequency disentanglement~\cite{tong2024lightweight}, and lightweight structural adaptation~\cite{fan2024adapting,su2024domain}, but still suffer from limited generalization under severe domain shifts. With the advent of visual foundation models~\cite{kirillov2023segment,oquab2023dinov2,rombach2022high,ravi2024sam2}, several works~\cite{yang2024tavp,he2024apseg,zhu2024unleashing,sun2024vrp,cuttano2025sansa,xu2025unlocking} leverage SAM’s zero-shot and promptable segmentation capability to tackle CD-FSS. However, most of these approaches still rely on training stages, introducing additional computational overhead~\cite{liu2023matcher,zhang2024bridge,herzog2024adapt}. In contrast, our work focuses on a training-free solution that no longer relies on domain-specific data for CD-FSS.

\noindent\textbf{SAM Adaptation for Downstream Tasks} is an under-explored problem, although SAM excels in dealing with in-domain images such as COCO~\cite{lin2014coco}. It is well recognized that SAM needs adaptations for downstream tasks, such as medical~\cite{chen2023samadapter,mazurowski2023samempiricalstudy,ma2024medsam,wu2025medicalsamadapter} or satellite images~\cite{li2025segearthov,yan2023ringmo,ding2024samcd,liu2025pointsam}, given the significant domain differences between SAM training data and these images. Early approaches involve the collection of large-scale labeled data in a wide spectrum of imaging modalities~\cite{ma2024medsam} for adaptation, which is expensive and could bring privacy concerns for practical use. Motivated by the strong capability of recent vision foundation models~\cite{oquab2023dinov2,kirillov2023segment,ravi2024sam2}, more studies explore efficient adaptation of SAM to these downstream tasks, with either lightweight architectural add-ons~\cite{chen2023samadapter,ke2023samhq} or few-shot samples from data perspective~\cite{xiao2024cat,xiao2024segment}. In light of few-shot SAM pipeline for in-domain images~\cite{zhang2023personalize,cuttano2025sansa}, we propose a strategy that reduces the need for large-scale or few-shot tuning, enabling cross-domain image segmentation using a few reference exemplars at inference time.
\begin{figure}[t]
    \centering
    % \vspace{-2mm}
    \includegraphics[width=0.92\linewidth]{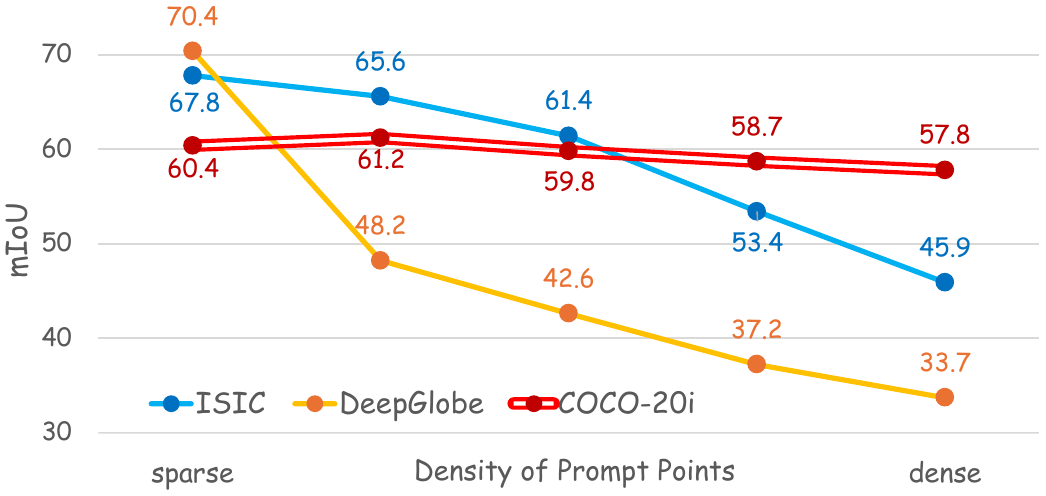}
    % \vspace{-1mm}
    \caption{Cross-domain and in-domain datasets exhibit varying sensitivities to prompt point density.}
    \label{fig:best_density}
    \vspace{-1mm}
\end{figure}

\section{Motivation and Preliminary}
\label{sec:motivation}

\subsection{Visual Foundation Models}
We briefly introduce two visual foundation models adopted in this paper, \textit{i.e.}, DINOv2~\cite{oquab2023dinov2} and SAM~\cite{kirillov2023segment}. \textbf{DINOv2}~\cite{oquab2023dinov2} is capable of extracting robust visual features for various downstream tasks and is particularly effective at capturing fine-grained local information~\cite{tong2024cambrian}. Consequently, prior works~\cite{liu2023matcher,zhang2024bridge} leverage DINOv2 to match dense points of target objects under few-shot settings. \textbf{SAM}~\cite{kirillov2023segment} is a promptable segmentation model that demonstrates strong generalization potential in both Cross-Domain Segmentation~\cite{xiao2024cat,xiao2024segment} and Few-Shot Segmentation~\cite{zhang2023personalize,liu2023matcher,zhang2024bridge} tasks. It is composed of three main components: an image encoder, a prompt encoder, and a mask decoder (as illustrated in Fig.~\ref{fig:motivation}(a)).

\subsection{Influence of Prompt Point Density}\label{ssec:influence}
To further validate our assumption in Sec.~\ref{sec:introduction} that prompt point density influences segmentation performance, we conduct comprehensive experiments on both in-domain (\textit{i.e.}, COCO-20i~\cite{nguyen2019feature}) and cross-domain (\textit{i.e.}, ISIC~\cite{codella2019skin,tschandl2018ham10000} and DeepGlobe~\cite{demir2018deepglobe}) datasets. Specifically, we place uniformly sampled points with varying densities on reference images and retain only those within the reference masks as prompt points, which are then fed into SAM for segmentation. As shown in Fig.~\ref{fig:best_density}, the in-domain dataset exhibits robust performance across different prompt point densities, whereas cross-domain datasets are more sensitive and suffer siginificant performance drop when dense points are used. Moreover, we observe that different categories achieve their best performance under different prompt point densities, indicating that a fixed density is suboptimal. For example, although most categories attain optimal performance with low-density points, some categories, such as \textit{nevus} in ISIC and \textit{cd}, \textit{pad}, and \textit{bucket} in COCO-20i, achieve their best performance with medium- or high-density points. These observations motivate the need for an adaptive mechanism to regulate prompt point density, leading us to propose CPS, which adaptively sparsifies prompt points, particularly for cross-domain scenarios.

\subsection{Accuracy of Matched Points}\label{ssec:point_acc}
In this paper, following previous work~\cite{zhang2024bridge}, we obtain prompt points on the target image via a dense matching strategy, which is described in detail in Sec.~\ref{ssec:dpm}. This strategy yields highly accurate prompt points on the target images, as analyzed in the appendix. These observations further suggest that the limitations of existing methods mainly lie in SAM’s point–image interaction, while the densely matched points provide a reliable initialization for our proposed method.

\subsection{Task Formulation}
Cross-Domain Few-Shot Segmentation (CD-FSS) aims to segment target objects in a cross-domain image conditioned on a few reference images. In this paper, we address the CD-FSS task in a training-free manner by leveraging the strong generalization capability of visual foundation models~\cite{kirillov2023segment,oquab2023dinov2}. Consequently, the task involves only an inference stage, and the problem setting can be simplified as follows. For each task, a reference image set $R=\{(I_i^r, M_i^r)\}_{i=1}^K$ is provided, which consists of $K$ reference images $\{I_i^r\}_{i=1}^K$ and corresponding segmentations masks $\{M_i^r\}_{i=1}^K$, along with a target image $I^t$. The goal is to predict an accurate segmentation mask $M^t$ of $I^t$, where the target object belongs to the same category as that specified in the reference image set $R$.
\begin{figure*}[t]
    \centering
    % \vspace{-3mm}
    \includegraphics[width=0.94\linewidth]{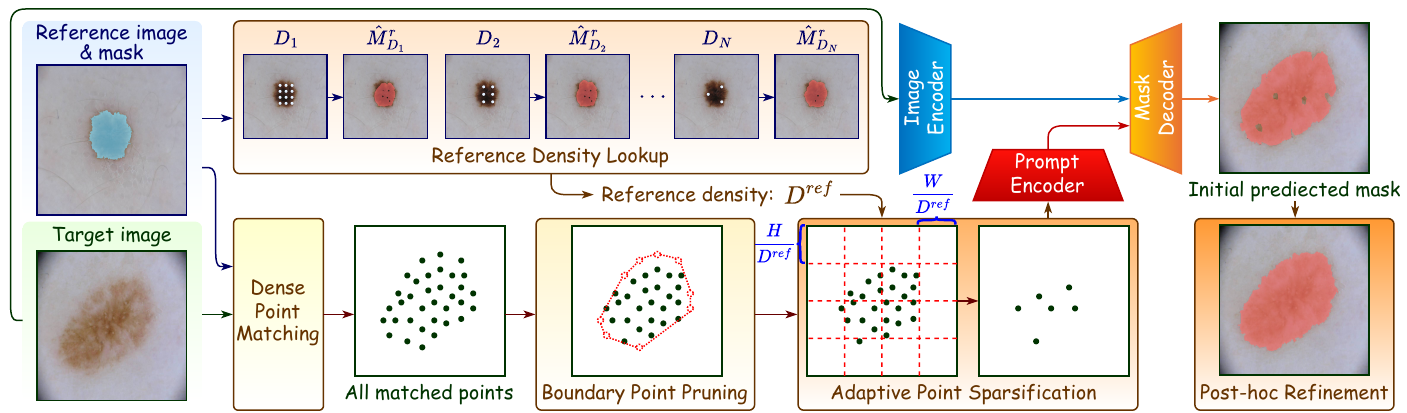}
    % \vspace{-1mm}
    \caption{Overview of the proposed Conditional Point Sparsification (CPS). CPS not only leverages the reference image to match candidate prompt points in the target image (Sec.~\ref{ssec:dpm}), but also exploits the reference image to determine an appropriate point density for subsequent sparsification (Sec.~\ref{ssec:cd}). The proposed modules, including boundary point pruning (Sec.~\ref{ssec:dpm}), adaptive point sparsification (Sec.~\ref{ssec:aps}), and post-hoc mask refinement (Sec.~\ref{ssec:pmr}), jointly contribute to producing accurate segmentation masks of the target image.}
    \label{fig:method}
    \vspace{-2mm}
\end{figure*}

\section{Method}\label{sec:method}
We describe our proposed Conditional Point Sparsification (CPS) in this section, which aims to provide suitable prompt points for SAM under the challenging Cross-Domain Few-Shot Segmentation (CD-FSS) task. The overall pipeline of CPS consists of four steps as illustrated in Fig.~\ref{fig:method}: \textit{(i)} dense point matching between the reference image $I^r$ and the target image $I^t$, followed by boundary point pruning to obtain an initial set of prompt points $\tilde{\mathbf{P}^t}$ for segmentation (Sec.~\ref{ssec:dpm}); \textit{(ii)} sparsification of the dense point set $\tilde{\mathbf{P}}^t$ on the target image to reduce ambiguity while preserving representative structural cues (Sec.~\ref{ssec:aps}); \textit{(iii)} reference point density lookup on the reference image $I^r$, which provides a conditional density signal to guide the sparsification process and is subsequently used to predict the initial segmentation mask $\hat{M}^t$ (Sec.~\ref{ssec:cd}); and \textit{(iv)} post-hoc refinement of the $\hat{M}^t$ and generation of the final mask $\tilde{M}^t$ (Sec.~\ref{ssec:pmr}).

\subsection{Dense Point Matching and Boundary Point Pruning}\label{ssec:dpm}

We first introduce the \textbf{Dense Point Matching} design for obtaining an initial set of prompt points, which serves as a proper initialization for the subsequent sparsification stage. The core idea is to exploit the correspondence between the reference image $I^r$ and the target image $I^t$ to identify candidate regions. To this end, we adopt the Positive-Negative Alignment (PNA) module proposed in prior work~\cite{zhang2024bridge} to perform support–query alignment. Specifically, PNA employs DINOv2~\cite{oquab2023dinov2} as the visual encoder to extract image features:
\vspace{-2mm}
\begin{equation}
F^r = \mathbf{DINOv2}(I^r), \quad F^t = \mathbf{DINOv2}(I^t),
\end{equation}
where $F^r \in \mathbb{R}^{h \times w}$ and $F^t \in \mathbb{R}^{h \times w}$ denote the feature maps of the reference image $I^r \in \mathbb{R}^{H \times W}$ and the target image $I^t \in \mathbb{R}^{H \times W}$, respectively. PNA then establishes dense correspondences between $F^r$ and $F^t$ by jointly utilizing foreground and background regions to consider global and local similarities, yielding a set of matched points on the target feature map $F^t$. Further implementation details are provided in the appendix. For simplicity, we denote the resulting matched point set as:
\vspace{-2mm}
\begin{equation}
\mathbf{p}^t = \{p_l^t\}_{l=1}^N = \mathrm{PNA}(F^r, F^t),
\end{equation}
where each matched point is parameterized as $p_l^t = (x_l^t, y_l^t)$ with $x_l^t \in \{0, \dots, w-1\}$ and $y_l^t \in \{0, \dots, h-1\}$. Here, $p_l^t$ denotes the $l$-th matched point on the target feature map $F^t$, and $\mathbf{p^t}$ represents a set of $N$ candidate target points that are highly likely to fall within the target region. Since each point $p_l^t \in \mathbf{p}^t$ is defined on the DINOv2 feature map of dimension $h \times w$, whereas SAM~\cite{kirillov2023segment} expects prompt points in the image space with input resolution $H \times W$, we project each point into the SAM input space. For each matched point $p_l^t = (x_l^t, y_l^t)$ on the feature map, its corresponding coordinate $P_l^t = (X_l^t, Y_l^t)$ in the SAM input space is computed as:
\vspace{-2mm}
\begin{equation}
X_l^t = \left(x_l^t + \frac{1}{2}\right) \cdot \frac{W}{w}, \quad Y_l^t = \left(y_l^t + \frac{1}{2}\right) \cdot \frac{H}{h},
\end{equation}
where the half-grid offset ensures that each projected point is aligned with the center of the corresponding image region. The resulting projected point set is:
\vspace{-2mm}
\begin{equation}
\mathbf{P}^t = \{ P_l^t \}_{l=1}^N,
\end{equation}
which lies in the same spatial domain as the SAM encoder and serves as the input for subsequent sparsification (refer to Sec.~\ref{ssec:point_acc}). However, this linear coordinate mapping may introduce spatial misalignment due to the resolution discrepancy between the feature map and the original image. Consequently, some projected points may be shifted toward boundary regions or even fall into background areas, thereby introducing unreliable prompts for SAM.

To mitigate these potential issues, we first propose a \textbf{Boundary Point Pruning} strategy.\footnote{For clarity, we describe the scenario with a single target object in this section. The extension to images containing multiple target objects is discussed in the appendix.} Specifically, given the mapped point set $\mathbf{P}^t$, we compute its convex hull~\cite{de2008computational} to approximate the coarse spatial extent of the target object. Formally, let:
\vspace{-2mm}
\begin{equation}
\mathcal{H}^t = \mathrm{ConvexHull}(\mathbf{P}^t)
\end{equation}
denote the convex hull of the mapped points, and let $\mathcal{V}(\mathcal{H}^t)$ represent the set of vertices of $\mathcal{H}^t$. Points lying on the convex hull boundary are more likely to be affected by spatial misalignment and background interference. Therefore, we remove these boundary points and retain only the interior points for subsequent processing:
\vspace{-2mm}
\begin{equation}
\tilde{\mathbf{P}^t} = \left\{ P^t_l \in \mathbf{P}^t \;\middle|\; P^t_l \notin \mathcal{V}(\mathcal{H}^t) \right\}.
\end{equation}
This operation effectively suppresses potentially noisy prompt points near object boundaries, yielding a more reliable prompt point set. Based on this refined set, we further perform point sparsification on $\tilde{\mathbf{P}^t}$ in Sec.~\ref{ssec:aps}.

\subsection{Adaptive Point Sparsification}\label{ssec:aps}

As discussed in Sec.~\ref{sec:introduction} and Sec.~\ref{sec:motivation}, although the dense point set $\tilde{\mathbf{P}}^t$ obtained in Sec.~\ref{ssec:dpm} provides a strong initialization, it is not suitable for CD-FSS. Specifically, significant discrepancies in visual patterns across domains prevent these dense points from serving as reliable prompts for cross-domain images. Moreover, excessively dense points tend to introduce ambiguity in target object segmentation. Therefore, it is necessary to sparsify the prompt points while preserving their structural representativeness.

A straightforward yet effective strategy is to sparsify points region-wise using a uniform grid. Specifically, we select an adaptive sparsification density $D^t$ (determined in Sec.~\ref{ssec:cd}) and partition the target image into a regular grid of $D^t \times D^t$ cells. Let:
\vspace{-2mm}
\begin{equation}\label{eq:density}
\Delta_h^t = \frac{H}{D^{t}}, \quad \Delta_w^t = \frac{W}{D^{t}}
\end{equation}
denote the cell height and width, respectively. Each grid cell is indexed by $(i,j)$, where $i,j \in \{0,1,\ldots,D^{t}-1\}$, and corresponds to the spatial region:
\vspace{-2mm}
\begin{equation}
\mathcal{G}_{i,j} = \left[i\Delta_h^t, (i+1)\Delta_h^t\right) \times
\left[j\Delta_w^t, (j+1)\Delta_w^t\right).
\end{equation}
To select a representative point from each cell, we first compute the global centroid of all remaining points $\tilde{\mathbf{P}}^t$:
\vspace{-2mm}
\begin{equation}
C = \frac{1}{|\tilde{\mathbf{P}}^{t}|} \sum_{\tilde{P}^t \in \tilde{\mathbf{P}}^{t}} \tilde{P}^t,
\end{equation}
which approximates the central location of the target object in the image space. For each grid cell $\mathcal{G}_{i,j}$, we collect all points from $\tilde{\mathbf{P}}^{t}$ that fall inside the cell:
\vspace{-2mm}
\begin{equation}
\tilde{\mathbf{P}}^{t}_{i,j} = \left\{\tilde{P}^t \in \tilde{\mathbf{P}}^{t}\;\middle|\; \tilde{P}^t \in \mathcal{G}_{i,j} \right\}.
\end{equation}
If $\tilde{\mathbf{P}}^{t}_{i,j}$ is non-empty, we retain only the point closest to the global centroid $C$:
\vspace{-2mm}
\begin{equation}
P^*_{i,j} = \arg\min_{\tilde{P}^t_{i,j} \in \tilde{\mathbf{P}}^{t}_{i,j}} \left\|\tilde{P}^t_{i,j} - C\right\|_2.
\end{equation}
The region-level sparsified point set is then obtained by aggregating the selected points from all grid cells:
\vspace{-2mm}
\begin{equation}
\hat{\mathbf{P}}^{t} = \bigcup_{i,j} P^{*}_{i,j}.
\end{equation}
By selecting points that are spatially distributed via grid partitioning, this strategy reduces the density of the $\tilde{\mathbf{P}}^t$ while preserving the structural cues of the target object.

\subsection{Conditional Design}\label{ssec:cd}

As discussed in Sec.~\ref{ssec:influence}, different datasets and object categories exhibit optimal segmentation performance under different prompt point densities. Inspired by prior works~\cite{liu2023matcher,zhang2024bridge} that utilize the reference image for point matching, we further exploit the reference image as a reliable exemplar to determine an appropriate point density $D^{ref}$ for sparsification in Sec.~\ref{ssec:aps}, rather than using a manually defined density.

First, we introduce the \textbf{Reference Density Lookup} design, which identifies a suitable point density from a set of candidate densities $\mathbf{D} = \{D_1, D_2, \ldots, D_N\}$. For each density $D_i \in \mathbf{D}$, we uniformly sample prompt points on the reference image $I^r \in \mathbb{R}^{H \times W}$ using a regular grid. The sampling interval along the height and width is computed as:
\vspace{-2mm}
\begin{equation}
\Delta_h^r = \frac{H}{D_i}, \quad \Delta_w^r = \frac{W}{D_i}.
\end{equation}
The resulting set of grid points is:
\vspace{-2mm}
\begin{equation}
\mathbf{P}^r_{D_i} = \left\{ \left(m \Delta_h^r, \, n \Delta_w^r \right) \;\middle|\; m,n \in \{0,1,\ldots,D_i\} \right\}.
\end{equation}
We then retain only the points that fall within the foreground region, where $M^r(\mathbf{\cdot}) = 1$. Formally, the filtered point set is:
\vspace{-2mm}
\begin{equation}
\tilde{\mathbf{P}}^r_{D_i} = \left\{ P^r_{D_i} \in \mathbf{P}^r_{D_i} \;\middle|\; M^r(P^r_{D_i}) = 1 \right\}.
\end{equation}
These retained points $\tilde{\mathbf{P}}^r_{D_i}$ are then used as prompts for SAM to predict a segmentation mask on the reference image:
\vspace{-2mm}
\begin{equation}
\hat{M}^r_{D_i} = \mathbf{SAM}\!\left(I^r, \tilde{\mathbf{P}}^r_{D_i}\right).
\end{equation}
Next, we evaluate the quality of each density $D_i$ using the Intersection over Union (IoU) between the predicted mask $\hat{M}^r_{D_i}$ and the ground-truth mask $M^r$:
\vspace{-2mm}
\begin{equation}
\mathrm{IoU}(D_i) = \frac{\left| \hat{M}^r_{D_i} \cap M^r \right|}{\left| \hat{M}^r_{D_i} \cup M^r \right|}.
\end{equation}
Finally, we select the density $D_i$ that yields the highest mIoU as the reference density $D^{ref}$:
\vspace{-2mm}
\begin{equation}
D^{ref} = \arg\max_{D_i \in \mathbf{D}} \mathrm{mIoU}(D_i).
\end{equation}
The selected conditional density $D^{ref}$ then replaces the $D^t$ in Eq.~\ref{eq:density}, enabling \textbf{Adaptive Point Sparsification} of $\tilde{\mathbf{P}}^t$ on the target image as described in Sec.~\ref{ssec:aps}, resulting in the sparsified point set $\hat{\mathbf{P}}^{t}$. The $\hat{\mathbf{P}}^{t}$ is then used to generate an initial segmentation mask for $I^t$ via SAM:
\vspace{-2mm}
\begin{equation}
\hat{M}^t = \mathbf{SAM}\left(I^t, \hat{\mathbf{P}}^{t}\right).
\end{equation}

\begin{table*}[t]
    \renewcommand\arraystretch{1.14}
    \centering
    % \vspace{-2mm}
    \caption{Quantitative comparisons between the proposed CPS and existing methods over four widely adopted Cross-Domain Few-Shot Segmentation benchmarks (in mIoU (\%)). The best mIoU number under each setup is highlighted by \textbf{bold} font.}
    \vspace{-1mm}
    \resizebox{\linewidth}{!}{
    \setlength{\tabcolsep}{20pt}
    \begin{tabular}{l||cc|cc|cc|cc}
        \toprule[1pt]
        \multicolumn{1}{c||}{\multirow{2}*{Method}}& \multicolumn{2}{c|}{ISIC}& \multicolumn{2}{c|}{Chest X-Ray}& \multicolumn{2}{c|}{DeepGlobe}& \multicolumn{2}{c}{SUIM}\\\cline{2-9}
        ~& 1-shot& 5-shot& 1-shot& 5-shot& 1-shot& 5-shot& 1-shot& 5-shot\\\hline\hline
        \multicolumn{9}{c}{\textit{\textbf{Training-required Methods}}}\\\hline
        RPMMs& 18.0& 20.0& 30.1& 30.8& 13.0& 13.5& -& -\\
        PGNet& 21.9& 21.3& 34.0& 28.0& 10.7& 12.4& -& -\\
        RePRI& 23.3& 26.2& 65.1& 65.5& 25.0& 27.4& -& -\\
        PFENet& 23.5& 23.8& 27.2& 27.6& 16.9& 18.0& -& -\\
        PANet& 25.3& 34.0& 57.8& 69.3& 36.6& 45.3& -& -\\
        CaNet& 25.2& 28.2& 28.4& 28.6& 22.3& 23.1& -& -\\
        AMP& 28.4& 30.4& 51.2& 53.0& 37.6& 40.6& -& -\\
        HSNet& 31.2& 35.1& 51.9& 54.4& 29.7& 35.1& 28.8& -\\
        PATNet& 41.2& 53.6& 66.6& 70.2& 37.9& 43.0& 32.1& 40.2\\
        ABCDFSS& 45.7& 53.3& 79.8& 81.4& 42.6& 49.0& 35.1& 41.3\\
        SSP& 48.6& 65.4& 72.6& 73.0& 41.3& 54.2& -& -\\\hline\hline

        \multicolumn{9}{c}{\textit{\textbf{SAM-based Training-free Methods}}}\\\hline
        PerSAM& 23.9& -& -& -& 31.4& -& -& -\\
        Matcher& 38.6& 35.0& 66.8& 66.3& 48.1& 50.9& 37.9& 41.4\\
        GF-SAM& 48.7& 55.2& 47.6& 47.4& 49.5& 57.7& 38.3& 42.9\\
        \cellcolor{mycolor}CPS& 
        \cellcolor{mycolor}\textbf{51.3}& \cellcolor{mycolor}\textbf{56.6}& \cellcolor{mycolor}\textbf{70.0}& \cellcolor{mycolor}\textbf{70.2}& \cellcolor{mycolor}46.0& \cellcolor{mycolor}49.8& \cellcolor{mycolor}\textbf{42.8}& \cellcolor{mycolor}\textbf{48.8}\\
        \bottomrule[1pt]
    \end{tabular}
    }
    % \vspace{-1mm}
    \label{tab:cdfss}
\end{table*}

\subsection{Post-hoc Mask Refinement}\label{ssec:pmr}

As shown in Fig.~\ref{fig:motivation}, the activation heatmap induced by prompt points may exhibit discontinuous patterns. Such discontinuities can result in unexpected holes within the target region or abrupt changes along object boundaries when SAM applies a fixed threshold~\cite{kirillov2023segment,liu2023matcher,zhang2024bridge} to generate the segmentation mask. To address this issue, we propose a post-hoc refinement strategy to regularize the initial predicted mask.

Specifically, we employ morphological opening and closing operations~\cite{serra1982image} to suppress isolated activations and fill small holes, thereby enforcing spatial continuity and improving boundary smoothness. Formally, $\hat{M}^t \in \mathbb{R}^{H \times W}$ denotes the initial predicted mask of the target image from Sec.~\ref{ssec:aps}, and let $K$ be a morphological structuring element. The refined prediction map $\tilde{M}^t$ is obtained by sequentially applying opening and closing operations:
\vspace{-2mm}
\begin{equation}
\tilde{M}^t = (\hat{M}^t \circ K) \bullet K,
\end{equation}
where $\circ$ and $\bullet$ represent morphological opening and closing, respectively. The opening operation removes small isolated activations, while the closing operation fills small gaps and holes, resulting in a more spatially coherent predicted mask.

\section{Experiment}\label{sec:experiment}
\subsection{Datasets}
We conduct extensive experiments on four cross-domain datasets spanning medical images~\cite{codella2019skin,tschandl2018ham10000,candemir2013lung,jaeger2013automatic}, satellite images~\cite{demir2018deepglobe}, and underwater scenes~\cite{islam2020semantic}. \textbf{ISIC2018}~\cite{codella2019skin,tschandl2018ham10000} consists of images for skin lesion analysis and covers three types of skin lesions. \textbf{Chest X-Ray}~\cite{candemir2013lung,jaeger2013automatic} is collected for tuberculosis screening. The grayscale nature of this dataset further increases the diversity of evaluation domains. \textbf{DeepGlobe}~\cite{demir2018deepglobe} is a satellite image dataset containing six terrain categories, including urban, agriculture, rangeland, forest, water, and barren areas. Following PATNet~\cite{lei2022cross}, we split the original images into smaller patches for the Cross-Domain Few-Shot Segmentation (CD-FSS) setting. \textbf{SUIM}~\cite{islam2020semantic} is an underwater image dataset comprising seven object categories, including fish, plants, divers, robots, ruins, and rocks.

\subsection{Implementation Details}
Following the experimental settings of Matcher~\cite{liu2023matcher} and GF-SAM~\cite{xu2024hybrid} for fair comparison, we adopt DINOv2~\cite{oquab2023dinov2} with a ViT-L/14 backbone~\cite{dosovitskiy2020image} to perform dense point matching and generate candidate prompt points. These points are refined by the proposed Boundary Point Pruning and Adaptive Point Sparsification modules. The remaining prompt points, together with the target image, are then fed into SAM~\cite{kirillov2023segment} with a ViT-H backbone~\cite{dosovitskiy2020image} to produce segmentation masks. Following Matcher and GF-SAM, the input image resolution is set to $518\times518$ for DINOv2 and $1024\times1024$ for SAM. We evaluate the segmentation performance using the mean Intersection over Union (mIoU) metric. All experiments are conducted on a single NVIDIA RTX 4090 GPU.

\begin{figure}[t]
    \centering
    % \vspace{-2mm}
    \includegraphics[width=0.99\linewidth]{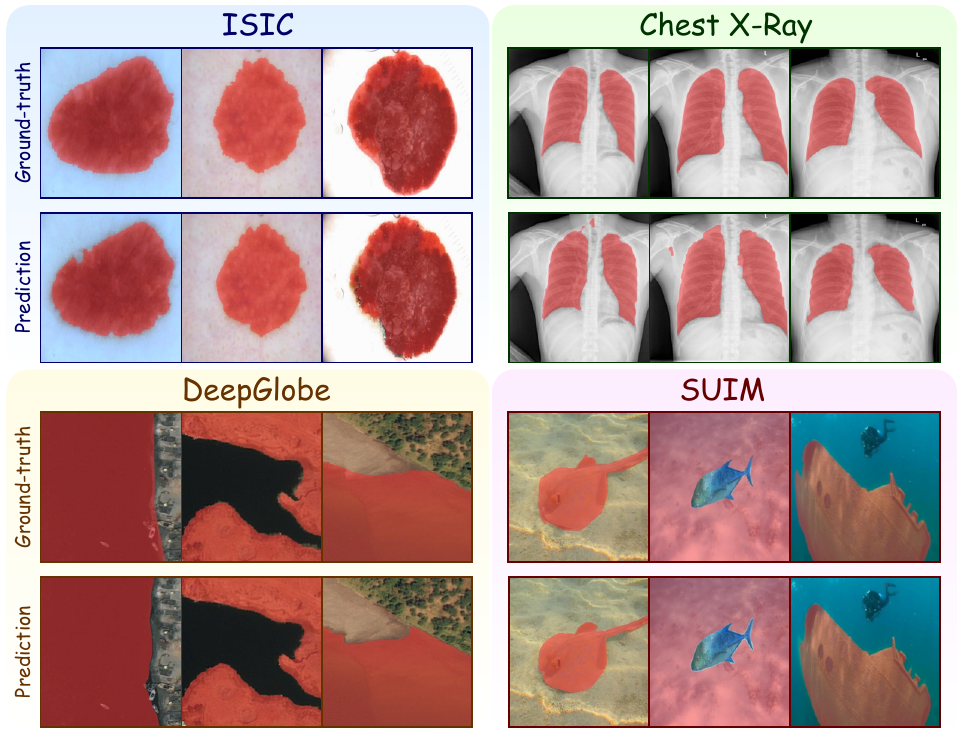}
    \vspace{-1mm}
    \caption{Qualitative segmentation results (red mask) of CPS on samples from four Cross-Domain Few-Shot Segmentation datasets. More examples are provided in the appendix.}
    \label{fig:cdfss_vis}
    \vspace{-2mm}
\end{figure}

\subsection{Comparison of Performance on CD-FSS Datasets}
We evaluate our proposed Conditional Point Sparsification (CPS) on four representative CD-FSS datasets and compare it with two groups of state-of-the-art methods in Tab.~\ref{tab:cdfss}. Overall, CPS consistently achieves superior performance across most subsets, often by a large margin. We first compare CPS with traditional CD-FSS methods~\cite{yang2020prototype,zhang2019pyramid,boudiaf2021few,tian2020prior,wang2019panet,zhang2019canet,siam2019amp,min2021hypercorrelation,lei2022cross,herzog2024adapt,fan2022self}, which are typically built upon relatively small backbones (\textit{e.g.}, ResNet-50~\cite{he2016deep}) and rely on task-specific training and fine-tuning. Despite being completely training-free, CPS demonstrates competitive or superior performance compared to these training-dependent approaches. Due to the high computational cost and limited cross-domain generalization of training-based pipelines, recent works have shifted toward either adaptation-only strategies~\cite{herzog2024adapt} or training-free methods that leverage the strong generalization ability of SAM~\cite{kirillov2023segment}~\cite{zhang2023personalize,liu2023matcher,zhang2024bridge}. We further benchmark CPS against these training-free SAM-based methods and observe consistent improvements, validating the effectiveness of our proposed point sparsification and refinement strategies.

From a performance perspective, MPA consistently delivers strong results across diverse domains, including medical imaging, remote sensing, and underwater scenes, demonstrating its robustness and cross-domain generalization capability. The medical images in ISIC~\cite{codella2019skin,tschandl2018ham10000} and Chest X-Ray~\cite{candemir2013lung,jaeger2013automatic} exhibit relatively clean backgrounds, with target regions occupying a large portion of the image. As analyzed in Fig.~\ref{fig:tsne}, the target regions demonstrate high intra-mask feature consistency, under which our point sparsification strategy performs particularly well. Specifically, it surpasses state-of-the-art SAM-based methods by 2.6\% and 3.2\% on ISIC and Chest X-Ray, respectively. For the underwater images in SUIM~\cite{islam2020semantic}, the background also contains multiple semantic objects, leading to a more complex and cluttered visual context. Nevertheless, CPS still achieves a new state-of-the-art performance, outperforming existing methods by 4.5\% under the 1-shot setup. DeepGlobe~\cite{demir2018deepglobe} is characterized by an aerial viewpoint and a highly complex background. Although its intra-mask feature consistency is higher than that of medical images (see Fig.~\ref{fig:tsne}), CPS still achieves competitive performance on this dataset. Moreover, it is worth noting that CPS consistently achieves better performance in the 5-shot setting than in the 1-shot setting, indicating that this training-free approach can leverage more informative prompts when additional support images are provided. In addition, we conduct qualitative comparisons on multiple CD-FSS datasets to further demonstrate the effectiveness of CPS. As shown in Fig.~\ref{fig:cdfss_vis}, CPS produces accurate and coherent segmentation masks.

From an efficiency perspective, CPS not only outperforms methods that require training or fine-tuning~\cite{lei2022cross,herzog2024adapt}, but also demonstrates higher efficiency than some SAM-based approaches~\cite{liu2023matcher}. More details are provided in the appendix.

\begin{figure}[t]
    \centering
    % \vspace{-2mm}
    \includegraphics[width=0.98\linewidth]{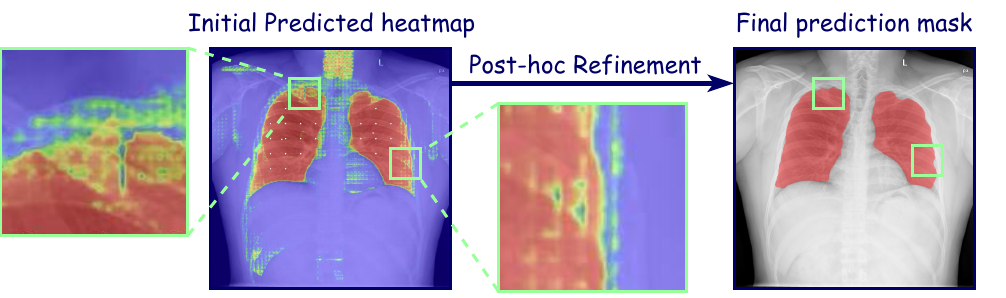}
    \vspace{-1mm}
    \caption{Illustrative example of Post-hoc Mask Refinement.}
    \label{fig:refine}
    \vspace{-1mm}
\end{figure}

\subsection{Discussion}\label{ssec:discussion}

% \noindent\textbf{Illustration of the CPS Pipeline.} As illustrated in Fig.~\ref{fig:step}, CPS first sparsifies the dense matched points and then utilizes the sparsified points as prompts to generate segmentation predictions (Sec.~\ref{sec:xxx}), followed by a refinement stage (Sec.~\ref{ssec:pmr}). The refinement process effectively improves the final predictions, particularly by enforcing spatial continuity and enhancing boundary smoothness.

\begin{table}[t]
    \renewcommand\arraystretch{1.14}
    \centering
    \vspace{+2mm}
    \caption{Ablation studies for technical designs of CPS. The Conditional Point Sparsification, Boundary Point Pruning, and Post-hoc Mask Refinement designs show significant effectiveness.}
    \vspace{-1mm}
    \resizebox{\linewidth}{!}{
    \setlength{\tabcolsep}{16pt}
    \begin{tabular}{l||cc}
        \toprule[1pt]
        Incorporated deisgn & ISIC & SUIM\\\hline\hline
        Baseline& 39.3& 34.2\\
        $+$Adaptive Point Sparsification& 47.2& 40.6\\
        $+$Boundary Point Pruning& 47.7& 41.2\\
        \cellcolor{mycolor}$+$Post-hoc Mask Refinement& \cellcolor{mycolor}\textbf{51.3} & \cellcolor{mycolor}\textbf{42.8}\\
        \bottomrule[1pt]
    \end{tabular}
    }
    \vspace{-1mm}
    \label{tab:ablation}
\end{table}

\begin{table*}[t]
    \renewcommand\arraystretch{1.12}
    \centering
    % \vspace{-2mm}
    \caption{Quantitative comparisons between the proposed CPS and existing methods over two widely adopted natural image Few-Shot Segmentation benchmarks (in mIoU (\%)). The best mIoU number under each setup is highlighted by \textbf{bold} font.}
    \vspace{-1mm}
    \resizebox{\linewidth}{!}{
    \setlength{\tabcolsep}{10pt}
    \begin{tabular}{l||ccccc|ccccc||l||cc}
        \toprule[1pt]
        \multicolumn{1}{c||}{\multirow{3}*{Method}}& \multicolumn{10}{c||}{COCO-20i}& \multicolumn{1}{c||}{\multirow{3}*{Method}} & \multicolumn{2}{c}{FSS-1000} \\\cline{2-11}\cline{13-14}
        ~& \multicolumn{5}{c|}{1-shot}& \multicolumn{5}{c||}{5-shot}& ~& \multirow{2}*{1-shot} & \multirow{2}*{5-shot} \\\cline{2-11}
        ~& $20^{0}$& $20^{1}$& $20^{2}$& $20^{3}$& Avg.& $20^{0}$& $20^{1}$& $20^{2}$& $20^{3}$& Avg.& ~& ~& ~\\\hline\hline
        % PFENet& 36.5& 38.6& 35.0& 33.8& 35.8& 36.5& 43.3& 38.0& 38.4& 39.0& PFENet& 70.9& 70.5\\
        SSP& 35.5& 39.6& 37.9& 36.7& 37.4& 40.6& 47.0& 45.1& 43.9& 44.1& RePRI& 71.0& 74.2\\
        CyCTR& 38.9& 43.0& 39.6& 39.8& 40.3& 41.1& 48.9& 45.2& 47.0& 45.6& ABCDFSS& 74.6& 76.2\\
        HSNet& 37.2& 44.1& 42.4& 41.3& 41.2& 45.9& 53.0& 51.8& 47.1& 49.5& SSP& 77.0& 79.4\\
        BAM& 43.4& 50.6& 47.5& 43.4& 46.2& 49.3& 54.2& 51.6& 49.6& 51.2& HSNet& 77.5& 81.0\\
        PAM& 44.1& 55.0& 46.5& 48.5& 48.5& 48.1& 60.8& 54.8& 51.9& 53.9& PATNet& 78.6& 81.2\\
        AENet& 43.1& \textbf{56.0}& \textbf{50.3}& 48.4& 49.4& 51.7& \textbf{61.9}& \textbf{57.9}& 55.3& 56.7& DR-Adapter& 79.1& 80.4\\
        HDMNet& 44.8& 54.9& 50.0& 48.7& \textbf{49.6}& 50.9& 60.2& 55.0& 55.3& 55.3& IFA& 80.1& 82.4\\
        PerSAM& -& -& -& -& 23.0& -& -& -& -& -& PerSAM& 71.2& - \\
        PerSAM-F& -& -& -& -& 23.5& -& -& -& -& -& PerSAM-F& 75.6& - \\
        \cellcolor{mycolor}CPS& \cellcolor{mycolor}\textbf{46.3}& \cellcolor{mycolor}52.6& \cellcolor{mycolor}48.2& \cellcolor{mycolor}\textbf{49.6}& \cellcolor{mycolor}49.2& \cellcolor{mycolor}\textbf{52.5}& \cellcolor{mycolor}56.2& \cellcolor{mycolor}53.1& \cellcolor{mycolor}\textbf{65.7}& \cellcolor{mycolor}\textbf{56.9}& \cellcolor{mycolor}\textbf{CPS}& \cellcolor{mycolor}\textbf{81.7}& \cellcolor{mycolor}\textbf{82.8}\\
        
        \bottomrule[1pt]
    \end{tabular}
    }
    \vspace{-1mm}
    \label{tab:fss}
\end{table*}

\noindent\textbf{Technical design ablation.}
We conduct ablation studies to evaluate the effectiveness of the proposed Boundary Point Pruning (Sec.~\ref{ssec:dpm}), Adaptive Point Sparsification (Sec.\ref{ssec:aps} and Sec.~\ref{ssec:cd}), and Post-hoc Mask Refinement (Sec.~\ref{ssec:pmr}) modules. As a baseline, we directly use all densely matched points obtained in Sec.~\ref{ssec:dpm} as input prompts for SAM. As reported in Tab.~\ref{tab:ablation}, this baseline yields unsatisfactory performance, indicating that although the matched points are relatively accurate (see Sec.~\ref{ssec:point_acc}), directly using dense points is not suitable for CD-FSS tasks. We then incorporate the core design, \textbf{Adaptive Point Sparsification}, which adaptively reduces the point density on the target image conditioned on the reference density derived from the reference image. This results in substantial performance improvements of 7.9\% and 6.4\% on the two evaluated datasets. Next, we further introduce \textbf{Boundary Point Pruning} to remove points that are more likely to fall into background regions, leading to additional gains of 0.5\% and 0.6\%, respectively. Finally, applying the proposed \textbf{Post-hoc Mask Refinement} module brings further improvements of 3.6\% and 1.6\% on the two datasets. These results demonstrate that each component contributes positively to the overall performance, and their combination yields the best results.

\begin{table}[t]
    \renewcommand\arraystretch{1.2}
    \centering
    % \vspace{-2mm}
    \caption{Effectiveness of Boundary Point Pruning in improving point accuracy. The accuracy represents the ratio of points that fall within the ground-truth masks of the target images.}
    \vspace{-1mm}
    \resizebox{\linewidth}{!}{
    \setlength{\tabcolsep}{7pt}
    \begin{tabular}{l||cc}
        \toprule[1pt]
        Setup & ISIC & Chest X-Ray\\\hline\hline
        Accuracy of all matched points & 77.2 & 94.7\\
        \cellcolor{mycolor}Accuracy after Boundary Point Pruning & \cellcolor{mycolor}\textbf{81.4} & \cellcolor{mycolor}\textbf{96.9}\\
        \bottomrule[1pt]
    \end{tabular}
    }
    \vspace{-1mm}
    \label{tab:boundary_prune}
\end{table}

\noindent\textbf{Effectiveness of Boundary Point Pruning.} As shown in Tab.~\ref{tab:boundary_prune}, Boundary Point Pruning improves the accuracy of matched points on two datasets. This indicates that this strategy filters out some inaccurately matched points, thereby providing more reliable prompts for segmentation.

\noindent\textbf{Illustrative example of Post-hoc Mask Refinement.} As shown in Fig.~\ref{fig:refine}, the refinement process effectively improves the final predictions, particularly by enforcing spatial continuity and enhancing boundary smoothness.

\noindent\textbf{Point sparsification strategies.}
We compare a fixed density sparsification strategy with the proposed Adaptive Point Sparsification in Tab.~\ref{tab:sparse_comp}. Specifically, the fixed density strategy adopts a manually defined point density as guidance, instead of adaptively selecting the density conditioned on the reference image. As a result, it achieves only limited improvements over the baseline and consistently underperforms our adaptive strategy. These results further validate the importance of adaptively determining an appropriate point density for each target image.

\subsection{Generalization Potential of CPS}
We also evaluate the proposed CPS on two in-domain datasets consisting of natural images that are similar to the training data of SAM~\cite{kirillov2023segment}. Specifically, we select \textbf{FSS-1000}\footnote{FSS-1000 is regarded as a ``cross-domain'' dataset in prior CD-FSS works, as they adopt ImageNet-pretrained backbones whose training distribution differs from FSS-1000. In contrast, for SAM-based methods~\cite{liu2023matcher,zhang2024bridge}, FSS-1000 is considered an ``in-domain'' dataset because it consists of natural images that largely fall within the training distribution of SAM.}~\cite{li2020fss}, which comprises natural images of everyday objects, and the more challenging \textbf{COCO-20i}~\cite{nguyen2019feature}, which covers 80 classes with complex and cluttered background. Both datasets contain tiny objects, enabling us to examine whether sparse points introduce negative effects. As shown in Tab.~\ref{tab:fss}, our proposed CPS achieves competitive performance on both datasets. More detailed analysis and qualitative segmentation results are provided in the appendix.

\begin{table}[t]
    \renewcommand\arraystretch{1.12}
    \centering
    % \vspace{-2mm}
    \caption{Comparison between our designed Adaptive Point Sparsification and fixed density sparsification.}
    \vspace{-1mm}
    \resizebox{\linewidth}{!}{
    \setlength{\tabcolsep}{12pt}
    \begin{tabular}{l||cc}
        \toprule[1pt]
        Incorporated deisgn & ISIC & SUIM\\\hline\hline
        Baseline& 39.3& 34.2\\
        $+$Fixed Density Sparsification ($D=4$)& 46.2& 36.1\\
        $+$Fixed Density Sparsification ($D=8$)& 44.5& 38.8\\
        $+$Fixed Density Sparsification ($D=12$)& 41.8& 38.2\\
        \cellcolor{mycolor}$+$Adaptive Point Sparsification& \cellcolor{mycolor}\textbf{47.2} & \cellcolor{mycolor}\textbf{40.6}\\
        \bottomrule[1pt]
    \end{tabular}
    }
    \vspace{-1mm}
    \label{tab:sparse_comp}
\end{table}

\section{Conclusion}
In this work, we investigate the challenges of applying dense prompt points in Cross-Domain Few-Shot Segmentation (CD-FSS), where large domain shifts significantly impair SAM’s mask prediction. We identify that point density plays a crucial role under such cross-domain conditions and propose Conditional Point Sparsification (CPS), a training-free method that adaptively guides SAM interactions based on reference exemplars. By leveraging ground-truth masks to appropriately sparsify dense matched points, CPS ensures more reliable and accurate segmentation across diverse domains. Extensive experiments on multiple CD-FSS datasets demonstrate that CPS outperforms existing training-free SAM-based approaches, validating the effectiveness of our designs in handling domain shifts. Overall, our study highlights the importance of carefully controlling prompt points in cross-domain scenarios and provides a practical solution for extending SAM’s capabilities to challenging CD-FSS segmentation tasks.

\bibliography{example_paper}

@String(CVPR= {IEEE Conf. Comput. Vis. Pattern Recog.})

@String(ICCV= {Int. Conf. Comput. Vis.})

@String(ECCV= {Eur. Conf. Comput. Vis.})

@String(BMVC= {Brit. Mach. Vis. Conf.})

@String(TIP  = {IEEE Trans. Image Process.})

@String(ICLR = {Int. Conf. Learn. Represent.})

@String(IJCAI = {IJCAI})

@String(AAAI = {AAAI})

@String(CVPRW= {IEEE Conf. Comput. Vis. Pattern Recog. Worksh.})

@String(CVPR  = {CVPR})

@String(ICCV  = {ICCV})

@String(ECCV  = {ECCV})

@String(BMVC  =	{BMVC})

@String(TIP   = {IEEE TIP})

@String(ICLR  = {ICLR})

@String(CVPRW= {CVPRW})

@inproceedings{demir2018deepglobe,
  title={Deepglobe 2018: A challenge to parse the earth through satellite images},
  author={Demir, Ilke and Koperski, Krzysztof and Lindenbaum, David and Pang, Guan and Huang, Jing and Basu, Saikat and Hughes, Forest and Tuia, Devis and Raskar, Ramesh},
  booktitle={CVPRW},
  year={2018}
}

@article{codella2019skin,
  title={Skin lesion analysis toward melanoma detection 2018: A challenge hosted by the international skin imaging collaboration (isic)},
  author={Codella, Noel and Rotemberg, Veronica and Tschandl, Philipp and Celebi, M Emre and Dusza, Stephen and Gutman, David and Helba, Brian and Kalloo, Aadi and Liopyris, Konstantinos and Marchetti, Michael and others},
  journal={arXiv preprint arXiv:1902.03368},
  year={2019}
}

@article{tschandl2018ham10000,
  title={The HAM10000 dataset, a large collection of multi-source dermatoscopic images of common pigmented skin lesions},
  author={Tschandl, Philipp and Rosendahl, Cliff and Kittler, Harald},
  journal={Scientific Data},
  year={2018}
}

@article{candemir2013lung,
  title={Lung segmentation in chest radiographs using anatomical atlases with nonrigid registration},
  author={Candemir, Sema and Jaeger, Stefan and Palaniappan, Kannappan and Musco, Jonathan P and Singh, Rahul K and Xue, Zhiyun and Karargyris, Alexandros and Antani, Sameer and Thoma, George and McDonald, Clement J},
  journal={TMI},
  year={2013}
}

@article{jaeger2013automatic,
  title={Automatic tuberculosis screening using chest radiographs},
  author={Jaeger, Stefan and Karargyris, Alexandros and Candemir, Sema and Folio, Les and Siegelman, Jenifer and Callaghan, Fiona and Xue, Zhiyun and Palaniappan, Kannappan and Singh, Rahul K and Antani, Sameer and others},
  journal={TMI},
  year={2013}
}

@inproceedings{li2020fss,
  title={Fss-1000: A 1000-class dataset for few-shot segmentation},
  author={Li, Xiang and Wei, Tianhan and Chen, Yau Pun and Tai, Yu-Wing and Tang, Chi-Keung},
  booktitle={CVPR},
  year={2020}
}

@inproceedings{islam2020semantic,
  title={Semantic segmentation of underwater imagery: Dataset and benchmark},
  author={Islam, Md Jahidul and Edge, Chelsey and Xiao, Yuyang and Luo, Peigen and Mehtaz, Muntaqim and Morse, Christopher and Enan, Sadman Sakib and Sattar, Junaed},
  booktitle={IROS},
  year={2020}
}

@inproceedings{liu2023matcher,
  title={Matcher: Segment anything with one shot using all-purpose feature matching},
  author={Liu, Yang and Zhu, Muzhi and Li, Hengtao and Chen, Hao and Wang, Xinlong and Shen, Chunhua},
  booktitle={ICLR},
  year={2023}
}

@inproceedings{zhang2024bridge,
  title={Bridge the points: Graph-based few-shot segment anything semantically},
  author={Zhang, Anqi and Gao, Guangyu and Jiao, Jianbo and Liu, Chi and Wei, Yunchao},
  booktitle={NeurIPS},
  year={2024}
}

@inproceedings{sun2024vrp,
  title={VRP-SAM: SAM with visual reference prompt},
  author={Sun, Yanpeng and Chen, Jiahui and Zhang, Shan and Zhang, Xinyu and Chen, Qiang and Zhang, Gang and Ding, Errui and Wang, Jingdong and Li, Zechao},
  booktitle={CVPR},
  year={2024}
}

@inproceedings{
    cuttano2025sansa,
    title={SANSA: Unleashing the Hidden Semantics in {SAM}2 for Few-Shot Segmentation},
    author={Claudia Cuttano and Gabriele Trivigno and Giuseppe Averta and Carlo Masone},
    booktitle={NeurIPS},
    year={2025}
}

@inproceedings{xu2025unlocking,
  title={Unlocking the Power of SAM 2 for Few-Shot Segmentation},
  author={Xu, Qianxiong and Zhu, Lanyun and Liu, Xuanyi and Lin, Guosheng and Long, Cheng and Li, Ziyue and Zhao, Rui},
  booktitle={ICML},
  year={2025}
}

@inproceedings{siam2019amp,
  title={Amp: Adaptive masked proxies for few-shot segmentation},
  author={Siam, Mennatullah and Oreshkin, Boris N and Jagersand, Martin},
  booktitle={ICCV},
  year={2019}
}

@inproceedings{zhang2019pyramid,
  title={Pyramid graph networks with connection attentions for region-based one-shot semantic segmentation},
  author={Zhang, Chi and Lin, Guosheng and Liu, Fayao and Guo, Jiushuang and Wu, Qingyao and Yao, Rui},
  booktitle={ICCV},
  year={2019}
}

@inproceedings{wang2019panet,
  title={Panet: Few-shot image semantic segmentation with prototype alignment},
  author={Wang, Kaixin and Liew, Jun Hao and Zou, Yingtian and Zhou, Daquan and Feng, Jiashi},
  booktitle={ICCV},
  year={2019}
}

@inproceedings{zhang2019canet,
  title={Canet: Class-agnostic segmentation networks with iterative refinement and attentive few-shot learning},
  author={Zhang, Chi and Lin, Guosheng and Liu, Fayao and Yao, Rui and Shen, Chunhua},
  booktitle={CVPR},
  year={2019}
}

@inproceedings{yang2020prototype,
  title={Prototype mixture models for few-shot semantic segmentation},
  author={Yang, Boyu and Liu, Chang and Li, Bohao and Jiao, Jianbin and Ye, Qixiang},
  booktitle={ECCV},
  year={2020}
}

@article{tian2020prior,
  title={Prior guided feature enrichment network for few-shot segmentation},
  author={Tian, Zhuotao and Zhao, Hengshuang and Shu, Michelle and Yang, Zhicheng and Li, Ruiyu and Jia, Jiaya},
  journal={TPAMI},
  year={2020}
}

@inproceedings{boudiaf2021few,
  title={Few-shot segmentation without meta-learning: A good transductive inference is all you need?},
  author={Boudiaf, Malik and Kervadec, Hoel and Masud, Ziko Imtiaz and Piantanida, Pablo and Ben Ayed, Ismail and Dolz, Jose},
  booktitle={CVPR},
  year={2021}
}

@inproceedings{min2021hypercorrelation,
  title={Hypercorrelation squeeze for few-shot segmentation},
  author={Min, Juhong and Kang, Dahyun and Cho, Minsu},
  booktitle={ICCV},
  year={2021}
}

@inproceedings{lei2022cross,
  title={Cross-Domain Few-Shot Semantic Segmentation},
  author={Lei, Shuo and Zhang, Xuchao and He, Jianfeng and Chen, Fanglan and Du, Bowen and Lu, Chang-Tien},
  booktitle={ECCV},
  year={2022}
}

@inproceedings{chen2024pixel,
  title={Pixel Matching Network for Cross-Domain Few-Shot Segmentation},
  author={Chen, Hao and Dong, Yonghan and Lu, Zheming and Yu, Yunlong and Han, Jungong},
  booktitle={WACV},
  year={2024}
}

@inproceedings{nie2024cross,
  title={Cross-Domain Few-Shot Segmentation via Iterative Support-Query Correspondence Mining},
  author={Nie, Jiahao and Xing, Yun and Zhang, Gongjie and Yan, Pei and Xiao, Aoran and Tan, Yap-Peng and Kot, Alex C and Lu, Shijian},
  booktitle={CVPR},
  year={2024}
}

@inproceedings{su2024domain,
  title={Domain-Rectifying Adapter for Cross-Domain Few-Shot Segmentation},
  author={Su, Jiapeng and Fan, Qi and Pei, Wenjie and Lu, Guangming and Chen, Fanglin},
  booktitle={CVPR},
  year={2024}
}

@inproceedings{he2024apseg,
  title={APSeg: Auto-Prompt Network for Cross-Domain Few-Shot Semantic Segmentation},
  author={He, Weizhao and Zhang, Yang and Zhuo, Wei and Shen, Linlin and Yang, Jiaqi and Deng, Songhe and Sun, Liang},
  booktitle={CVPR},
  year={2024}
}

@inproceedings{herzog2024adapt,
  title={Adapt Before Comparison: A New Perspective on Cross-Domain Few-Shot Segmentation},
  author={Herzog, Jonas},
  booktitle={CVPR},
  year={2024}
}

@inproceedings{wang2022remember,
  title={Remember the Difference: Cross-Domain Few-Shot Semantic Segmentation via Meta-Memory Transfer},
  author={Wang, Wenjian and Duan, Lijuan and Wang, Yuxi and En, Qing and Fan, Junsong and Zhang, Zhaoxiang},
  booktitle={CVPR},
  year={2022}
}

@inproceedings{chen2024cross,
  title={Cross-Domain Few-Shot Semantic Segmentation via Doubly Matching Transformation},
  author={Chen, Jiayi and Quan, Rong and Qin, Jie},
  booktitle={IJCAI},
  year={2024}
}

@article{fan2023darnet,
  title={DARNet: Bridging Domain Gaps in Cross-Domain Few-Shot Segmentation with Dynamic Adaptation},
  author={Fan, Haoran and Fan, Qi and Pagnucco, Maurice and Song, Yang},
  journal={arXiv preprint arXiv:2312.04813},
  year={2023}
}

@inproceedings{huang2023restnet,
  title={Restnet: Boosting cross-domain few-shot segmentation with residual transformation network},
  author={Huang, Xinyang and Zhu, Chuang and Chen, Wenkai},
  booktitle={BMVC},
  year={2023}
}

@inproceedings{zhang2023personalize,
  title={Personalize segment anything model with one shot},
  author={Zhang, Renrui and Jiang, Zhengkai and Guo, Ziyu and Yan, Shilin and Pan, Junting and Ma, Xianzheng and Dong, Hao and Gao, Peng and Li, Hongsheng},
  booktitle={ICLR},
  year={2023}
}

@article{yang2024tavp,
  title={TAVP: Task-Adaptive Visual Prompt for Cross-domain Few-shot Segmentation},
  author={Yang, Jiaqi and Huang, Ye and He, Xiangjian and Shen, Linlin and Qiu, Guoping},
  journal={arXiv preprint arXiv:2409.05393},
  year={2024}
}

@article{fan2024adapting,
  title={Adapting In-Domain Few-Shot Segmentation to New Domains without Retraining},
  author={Fan, Qi and Liu, Kaiqi and Liu, Nian and Cholakkal, Hisham and Anwer, Rao Muhammad and Li, Wenbin and Gao, Yang},
  journal={arXiv preprint arXiv:2504.21414},
  year={2025}
}

@inproceedings{tong2024lightweight,
  title={Lightweight Frequency Masker for Cross-Domain Few-Shot Semantic Segmentation},
  author={Tong, Jintao and Zou, Yixiong and Li, Yuhua and Li, Ruixuan},
  booktitle={NeurIPS},
  year={2024}
}

@article{tong2025self,
  title={Self-Disentanglement and Re-Composition for Cross-Domain Few-Shot Segmentation},
  author={Tong, Jintao and Zou, Yixiong and Chen, Guangyao and Li, Yuhua and Li, Ruixuan},
  journal={arXiv preprint arXiv:2506.02677},
  year={2025}
}

@inproceedings{nguyen2019feature,
  title={Feature weighting and boosting for few-shot segmentation},
  author={Nguyen, Khoi and Todorovic, Sinisa},
  booktitle={ICCV},
  year={2019}
}

@inproceedings{fan2022self,
  title={Self-support few-shot semantic segmentation},
  author={Fan, Qi and Pei, Wenjie and Tai, Yu-Wing and Tang, Chi-Keung},
  booktitle={ECCV},
  year={2022}
}

@inproceedings{zhu2024unleashing,
  title={Unleashing the potential of the diffusion model in few-shot semantic segmentation},
  author={Zhu, Muzhi and Liu, Yang and Luo, Zekai and Jing, Chenchen and Chen, Hao and Xu, Guangkai and Wang, Xinlong and Shen, Chunhua},
  booktitle={NeurIPS},
  year={2024}
}

@inproceedings{xu2024hybrid,
  title={Hybrid mamba for few-shot segmentation},
  author={Xu, Qianxiong and Liu, Xuanyi and Zhu, Lanyun and Lin, Guosheng and Long, Cheng and Li, Ziyue and Zhao, Rui},
  booktitle={NeurIPS},
  year={2024}
}

@article{lang2023base,
  title={Base and meta: A new perspective on few-shot segmentation},
  author={Lang, Chunbo and Cheng, Gong and Tu, Binfei and Li, Chao and Han, Junwei},
  journal={TPAMI},
  year={2023}
}

@inproceedings{lu2021simpler,
  title={Simpler is better: Few-shot semantic segmentation with classifier weight transformer},
  author={Lu, Zhihe and He, Sen and Zhu, Xiatian and Zhang, Li and Song, Yi-Zhe and Xiang, Tao},
  booktitle={ICCV},
  year={2021}
}

@inproceedings{wu2024task,
  title={Task-adaptive prompted transformer for cross-domain few-shot learning},
  author={Wu, Jiamin and Liu, Xin and Yin, Xiaotian and Zhang, Tianzhu and Zhang, Yongdong},
  booktitle={AAAI},
  year={2024}
}

@inproceedings{fu2024cross,
  title={Cross-domain few-shot object detection via enhanced open-set object detector},
  author={Fu, Yuqian and Wang, Yu and Pan, Yixuan and Huai, Lian and Qiu, Xingyu and Shangguan, Zeyu and Liu, Tong and Fu, Yanwei and Van Gool, Luc and Jiang, Xingqun},
  booktitle={ECCV},
  year={2024}
}

@inproceedings{he2016deep,
  title={Deep residual learning for image recognition},
  author={He, Kaiming and Zhang, Xiangyu and Ren, Shaoqing and Sun, Jian},
  booktitle={CVPR},
  year={2016}
}

@inproceedings{deng2009imagenet,
  title={Imagenet: A large-scale hierarchical image database},
  author={Deng, Jia and Dong, Wei and Socher, Richard and Li, Li-Jia and Li, Kai and Fei-Fei, Li},
  booktitle={CVPR},
  year={2009}
}

@article{oquab2023dinov2,
  title={Dinov2: Learning robust visual features without supervision},
  author={Oquab, Maxime and Darcet, Timoth{\'e}e and Moutakanni, Th{\'e}o and Vo, Huy and Szafraniec, Marc and Khalidov, Vasil and Fernandez, Pierre and Haziza, Daniel and Massa, Francisco and El-Nouby, Alaaeldin and others},
  journal={arXiv preprint arXiv:2304.07193},
  year={2023}
}

@inproceedings{kirillov2023segment,
  title={Segment anything},
  author={Kirillov, Alexander and Mintun, Eric and Ravi, Nikhila and Mao, Hanzi and Rolland, Chloe and Gustafson, Laura and Xiao, Tete and Whitehead, Spencer and Berg, Alexander C and Lo, Wan-Yen and others},
  booktitle={ICCV},
  year={2023}
}

@article{ravi2024sam2,
  title={Sam 2: Segment anything in images and videos},
  author={Ravi, Nikhila and Gabeur, Valentin and Hu, Yuan-Ting and Hu, Ronghang and Ryali, Chaitanya and Ma, Tengyu and Khedr, Haitham and R{\"a}dle, Roman and Rolland, Chloe and Gustafson, Laura and others},
  journal={arXiv preprint arXiv:2408.00714},
  year={2024}
}

@inproceedings{tong2024cambrian,
  title={Cambrian-1: A fully open, vision-centric exploration of multimodal llms},
  author={Tong, Peter and Brown, Ellis and Wu, Penghao and Woo, Sanghyun and IYER, Adithya Jairam Vedagiri and Akula, Sai Charitha and Yang, Shusheng and Yang, Jihan and Middepogu, Manoj and Wang, Ziteng and others},
  booktitle={NeurIPS},
  year={2024}
}

@inproceedings{lin2014coco,
  title={Microsoft coco: Common objects in context},
  author={Lin, Tsung-Yi and Maire, Michael and Belongie, Serge and Hays, James and Perona, Pietro and Ramanan, Deva and Doll{\'a}r, Piotr and Zitnick, C Lawrence},
  booktitle={ECCV},
  organization={Springer},
  year={2014}
}

@article{maaten2008tsne,
  title={Visualizing data using t-SNE},
  author={Maaten, Laurens van der and Hinton, Geoffrey},
  journal={JMLR},
  year={2008}
}

@inproceedings{dosovitskiy2020image,
  title={An image is worth 16x16 words: Transformers for image recognition at scale},
  author={Dosovitskiy, Alexey},
  booktitle={ICLR},
  year={2020}
}

@inproceedings{rombach2022high,
  title={High-resolution image synthesis with latent diffusion models},
  author={Rombach, Robin and Blattmann, Andreas and Lorenz, Dominik and Esser, Patrick and Ommer, Bj{\"o}rn},
  booktitle={CVPR},
  year={2022}
}

@book{de2008computational,
  title={Computational geometry: algorithms and applications},
  author={De Berg, Mark and Cheong, Otfried and Van Kreveld, Marc and Overmars, Mark},
  year={2008},
  publisher={Springer}
}

@book{serra1982image,
  title={Image Analysis and Mathematical Morphology},
  author={Serra, Jean},
  year={1982}
}

@inproceedings{xiao2024cat,
  title={Cat-sam: Conditional tuning for few-shot adaptation of segment anything model},
  author={Xiao, Aoran and Xuan, Weihao and Qi, Heli and Xing, Yun and Ren, Ruijie and Zhang, Xiaoqin and Shao, Ling and Lu, Shijian},
  booktitle={ECCV},
  year={2024}
}

@article{xiao2024segment,
  title={Segment anything with multiple modalities},
  author={Xiao, Aoran and Xuan, Weihao and Qi, Heli and Xing, Yun and Yokoya, Naoto and Lu, Shijian},
  journal={arXiv preprint arXiv:2408.09085},
  year={2024}
}

@article{chen2023samadapter,
  title={SAM Fails to Segment Anything?--SAM-Adapter: Adapting SAM in Underperformed Scenes: Camouflage, Shadow, Medical Image Segmentation, and More},
  author={Chen, Tianrun and Zhu, Lanyun and Ding, Chaotao and Cao, Runlong and Wang, Yan and Li, Zejian and Sun, Lingyun and Mao, Papa and Zang, Ying},
  journal={arXiv preprint arXiv:2304.09148},
  year={2023}
}

@article{mazurowski2023samempiricalstudy,
  title={Segment anything model for medical image analysis: an experimental study},
  author={Mazurowski, Maciej A and Dong, Haoyu and Gu, Hanxue and Yang, Jichen and Konz, Nicholas and Zhang, Yixin},
  journal={Medical Image Analysis},
  year={2023}
}

@article{wu2025medicalsamadapter,
  title={Medical sam adapter: Adapting segment anything model for medical image segmentation},
  author={Wu, Junde and Wang, Ziyue and Hong, Mingxuan and Ji, Wei and Fu, Huazhu and Xu, Yanwu and Xu, Min and Jin, Yueming},
  journal={Medical image analysis},
  year={2025}
}

@article{ma2024medsam,
  title={Segment anything in medical images},
  author={Ma, Jun and He, Yuting and Li, Feifei and Han, Lin and You, Chenyu and Wang, Bo},
  journal={Nature Communications},
  year={2024}
}

@inproceedings{li2025segearthov,
  title={Segearth-ov: Towards training-free open-vocabulary segmentation for remote sensing images},
  author={Li, Kaiyu and Liu, Ruixun and Cao, Xiangyong and Bai, Xueru and Zhou, Feng and Meng, Deyu and Wang, Zhi},
  booktitle={CVPR},
  year={2025}
}

@article{yan2023ringmo,
  title={RingMo-SAM: A foundation model for segment anything in multimodal remote-sensing images},
  author={Yan, Zhiyuan and Li, Junxi and Li, Xuexue and Zhou, Ruixue and Zhang, Wenkai and Feng, Yingchao and Diao, Wenhui and Fu, Kun and Sun, Xian},
  journal={TGRS},
  year={2023}
}

@article{liu2025pointsam,
  title={Pointsam: Pointly-supervised segment anything model for remote sensing images},
  author={Liu, Nanqing and Xu, Xun and Su, Yongyi and Zhang, Haojie and Li, Heng-Chao},
  journal={TGRS},
  year={2025}
}

@article{ding2024samcd,
  title={Adapting segment anything model for change detection in VHR remote sensing images},
  author={Ding, Lei and Zhu, Kun and Peng, Daifeng and Tang, Hao and Yang, Kuiwu and Bruzzone, Lorenzo},
  journal={TGRS},
  year={2024}
}

@inproceedings{ke2023samhq,
  title={Segment anything in high quality},
  author={Ke, Lei and Ye, Mingqiao and Danelljan, Martin and Tai, Yu-Wing and Tang, Chi-Keung and Yu, Fisher and others},
  booktitle={NeurIPS},
  year={2023}
}

@inproceedings{fang2025sam2act,
  title={Sam2act: Integrating visual foundation model with a memory architecture for robotic manipulation},
  author={Fang, Haoquan and Grotz, Markus and Pumacay, Wilbert and Wang, Yi Ru and Fox, Dieter and Krishna, Ranjay and Duan, Jiafei},
  booktitle={ICML},
  year={2025}
}

@inproceedings{pan2025omnimanip,
  title={Omnimanip: Towards general robotic manipulation via object-centric interaction primitives as spatial constraints},
  author={Pan, Mingjie and Zhang, Jiyao and Wu, Tianshu and Zhao, Yinghao and Gao, Wenlong and Dong, Hao},
  booktitle={CVPR},
  year={2025}
}

@article{zhang2025alps,
  title={Alps: An auto-labeling and pre-training scheme for remote sensing segmentation with segment anything model},
  author={Zhang, Song and Wang, Qingzhong and Liu, Junyi and Xiong, Haoyi},
  journal={TIP},
  year={2025}
}

@article{wang2023sammed,
  title={Sammed: A medical image annotation framework based on large vision model},
  author={Wang, Chenglong and Li, Dexuan and Wang, Sucheng and Zhang, Chengxiu and Wang, Yida and Liu, Yun and Yang, Guang},
  journal={arXiv preprint arXiv:2307.05617},
  volume={3},
  year={2023}
}

@article{yang2025sam,
  title={SAM-guided Pseudo Label Enhancement for Multi-modal 3D Semantic Segmentation},
  author={Yang, Mingyu and Lu, Jitong and Kim, Hun-Seok},
  journal={arXiv preprint arXiv:2502.00960},
  year={2025}
}
\bibliographystyle{icml2026}

\end{document}